\documentclass[11pt,a4paper]{article}
\usepackage[hyperref]{acl2021}
\usepackage{times}
\usepackage{latexsym}
\usepackage{comment}

\usepackage{subfigure}
\usepackage{microtype}

\aclfinalcopy % Uncomment this line for the final submission
\usepackage{graphicx}
\graphicspath{ {./images/} }
\usepackage{amsmath}
\usepackage{array}
\usepackage{dirtytalk}
\usepackage[utf8]{inputenc}
\usepackage{enumitem}
\usepackage{xspace}
\usepackage{booktabs}

\newcommand{\ie}{\emph{i.e.,}\xspace}
\newcommand{\eg}{\emph{e.g.,}\xspace}

\newcommand{\dname}{DocOIE\xspace}
\newcommand{\mname}{DocIE\xspace}

\usepackage{multirow}

\usepackage{textcomp}
\newcommand{\textapprox}{\raisebox{0.5ex}{\texttildelow}}

\title{DocOIE: A Document-level Context-Aware Dataset for OpenIE}

\author{Kuicai Dong$^1$, Yilin Zhao$^1$, Aixin Sun$^1$, Jung-Jae Kim$^2$, Xiaoli Li$^{1,2}$ \\
$^1$
School of Computer Science and Engineering, Nanyang Technological University, Singapore\\
\texttt{\{kuicai001, zhao0320\}@e.ntu.edu.sg, \{axsun, xlli\}@ntu.edu.sg}\\
$^2$Institute for Infocomm Research, A*STAR, Singapore\\
\texttt{\{jjkim, xlli\}@i2r.a-star.edu.sg}}

\date{}

\begin{document}
\maketitle

%=============================================================================================
\begin{abstract}
Open Information Extraction (OpenIE) aims to extract structured relational tuples (subject, relation, object) from sentences and plays critical roles for many downstream NLP applications. Existing solutions perform extraction at \textit{sentence level}, without referring to any additional contextual information. In reality, however, a sentence typically exists as part of a document rather than standalone; we often need to access relevant contextual information around the sentence before we can accurately interpret it.
As there is no document-level context-aware OpenIE dataset available, we manually annotate 800 sentences from 80 documents in two domains (Healthcare and Transportation) to form a DocOIE dataset for evaluation.
In addition, we propose DocIE, a novel document-level context-aware OpenIE model. Our experimental results based on DocIE demonstrate that incorporating document-level context is helpful in improving OpenIE performance. Both DocOIE dataset and DocIE model are released for public~\footnote{https://github.com/daviddongkc/DocOIE}.

\end{abstract}
%=============================================================================================

%=============================================================================================
\section{Introduction}
\label{sec:intro}
%=============================================================================================

Open Information Extraction (OpenIE) has been a critical NLP task as it can extract structured relational tuples (\textit{subject, relation, object}) from unstructured text. The OpenIE system is fully domain-independent, and does not need input from users. It is also highly scalable that allows fast querying mechanism~\cite{yates2007textrunner}. Therefore, OpenIE has been successfully applied to a variety of downstream NLP tasks, such as knowledge base population~\cite{martinez2018openie, gashteovski2020aligning}, question answering~\cite{khot2017answering}, and summarization~\cite{fan2019using}.

\begin{figure}[th]
    \centering
    \includegraphics[width=0.95\columnwidth]{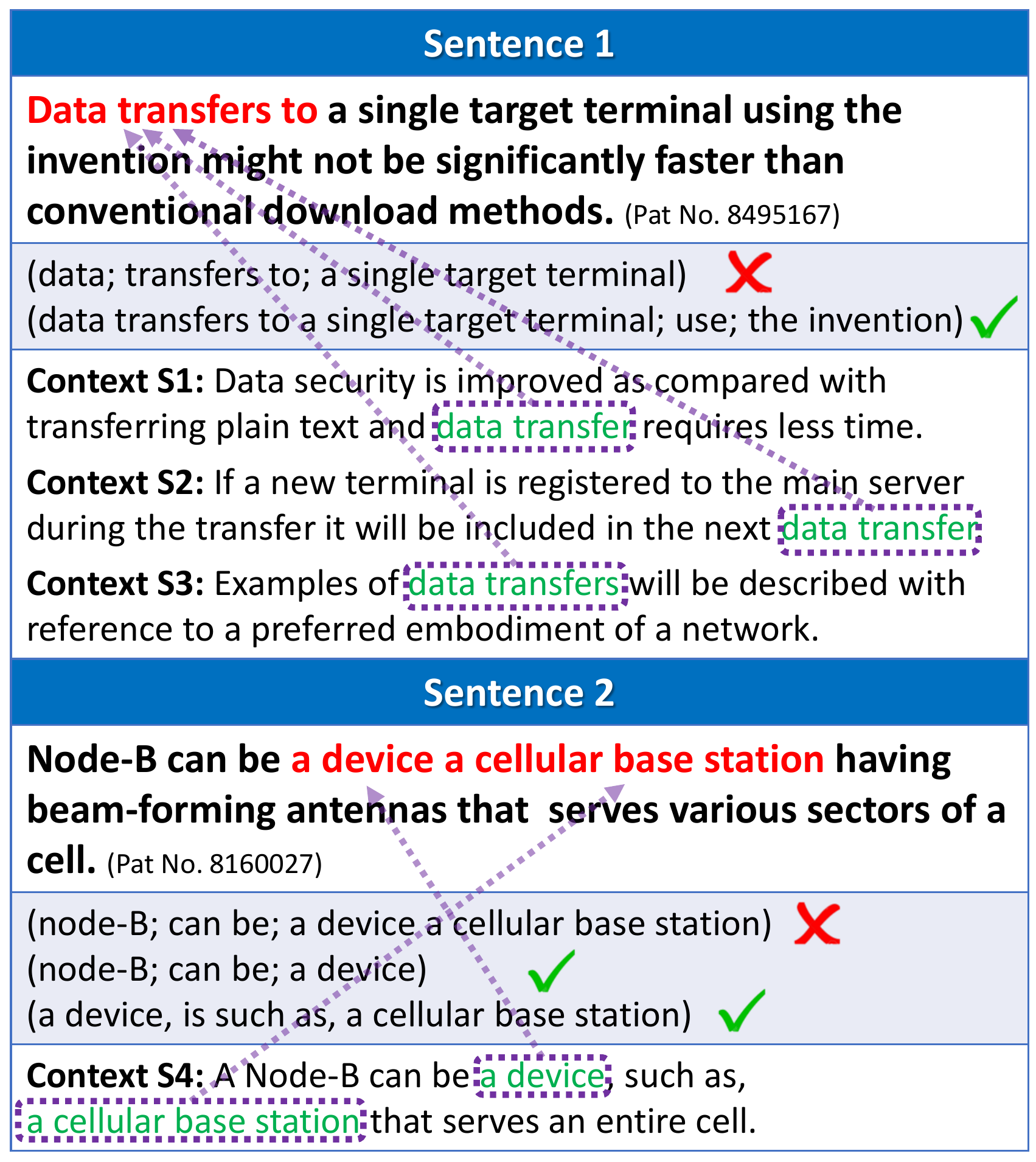}
    \vspace{-0.5em}
    \caption{Example sentences with ambiguity.}
    \vspace{-0.5em}
    \label{fig:example}
\end{figure}

Current OpenIE methods mainly focus on \textit{extracting tuples at the sentence-level}. However, in many NLP scenarios, sentences exist as part of a document rather than standalone. Given a document corpus, if we simply apply existing sentence-level OpenIE models to extract tuples, we could miss some useful and critical document-level contextual information, leading to unsatisfied results. We use the two example sentences in Fig.~\ref{fig:example} to illustrate two types of ambiguities. 

\paragraph{Part-of-speech Ambiguity.} The word \say{transfers} can be a verb or a noun. Accordingly, two tuples could be extracted from the first example sentence, listed in Fig.~\ref{fig:example}.
This ambiguity can be resolved by the main verb of the sentence \say{might not be}, which is far away from \say{transfers}, and thus is not considered by many existing OpenIE systems. However, context sentences S1, S2 and S3 in the document suggest that \say{data transfers} shall be considered as a noun phrase throughout this document. 

\paragraph{Syntactic Ambiguity.} The second example sentence does not have an explicit clue about the relationship between \say{a device} and \say{a cellular base station}.\footnote{The missing of a comma could be a typo in the patent document. Nevertheless, not all input sentences to OpenIE system are typo free in real applications.} Thus, existing OpenIE systems often fail to split them, but incorrectly extract the first tuple (node-B; can be; a device a cellular base station). However, context sentence S4 includes an explicit cue to the relationship between the two terms and may thus help split them.

To minimize the aforementioned ambiguities, it is clear that we should leverage document-level context. However, all existing OpenIE datasets are generated or annotated at sentence-level. These datasets include standalone sentences but not their context sentences. Hence they are not suitable for evaluating context-aware tuple extraction.

We annotate the first \textbf{Doc}ument-level context-aware \textbf{O}pen \textbf{I}nformation \textbf{E}xtraction (\textbf{\dname}) dataset. \dname consists of 800 expert annotated sentences from 80 documents; 10 sentences are randomly sampled for annotation from each of 80 documents. To the best of our knowledge, among all OpenIE datasets as of now, 
\dname  contains the largest number of \textit{expert-annotated} sentences. More importantly, \dname provides  document-level contexts, enabling OpenIE models to take relevant contexts for accurate tuple extraction.

Furthermore, to show that document-level context is useful for OpenIE task, we develop  the \textbf{Doc}ument-level context-aware Open \textbf{I}nformation \textbf{E}xtraction (\textbf{\mname}) model. \mname encodes a source sentence with its contextual information by using pre-trained BERT~\cite{devlin2018bert}. Because contextual sentences can be much longer than the source sentence, the syntactic/semantic information in source sentence might be dominated by that of the contexts. Our proposed \mname model differentiates the source sentence and its contexts by segment tags, and adds additional transformer encoder layers only for the source sentence. In summary, our contributions are threefold:
\setlist{nolistsep}
\begin{itemize}[noitemsep,leftmargin=5.5mm]
  \item We propose a new task in OpenIE to extract relational tuples with document-level contexts.
  \item We introduce \textbf{\dname}, \textit{an expert annotated dataset} for evaluating document-level OpenIE systems. \dname consists of 2,122 relational tuples in 800 expert-annotated sentences, with their document-level contexts.
  \item We present \textbf{\mname}, a novel neural OpenIE system that can leverage on document-level contexts for relational tuple extraction. 
\end{itemize}

%=============================================================================================
\section{Related Work}
\label{sec:related}
%=============================================================================================
\paragraph{OpenIE Datasets.} Since the OpenIE task was introduced by~\citet{yates2007textrunner}, the earlier systems were mainly evaluated by using a small number of sentences, without a standardized evaluation procedure~\cite{niklaus-etal-2018-survey}. OIE2016~\cite{stanovsky2016creating} is the first large-scale dataset constructed for OpenIE tasks and comes with a standard scoring framework. In OIE2016, the gold tuples are automatically generated from a QA-SRL dataset~\cite{he2015question} according to human crafted rules. Wire57~\cite{lechelle2018wire57} improves the scorer and manually annotates 57 sentences as a benchmark dataset. Considering that OIE2016 dataset is noisy, \citet{bhardwaj2019carb} provide a crowdsourcing dataset named CaRB. CaRB also has 50 expert annotated sentences and a sophisticated scoring framework.

As summarized in Table~\ref{tab:dataset_stats}, the number of expert-annotated sentences in these datasets remains small. Furthermore, the sentences in these datasets do not come with contextual information. In contrast, our \dname dataset consists of 800 expert-annotated sentences, and comes with the source documents for accurate sentence interpretation.

\begin{table}[t]
\centering
\begin{tabular}{l|rp{0.55in}l}
 \toprule
 Dataset & \#Sent. & Source & Annotation\\
 \midrule
 OIE2016 & 3,200& QA-SRL & Automatic \\
 Wire57 & 57 & Wikipedia Newswire & Expert \\
 CaRB & 1,282 & OIE2016 & Crowdsourcing\\
 CaRB & 50 & OIE2016 & Expert\\
 \midrule
 \dname & 800 & Patent & Expert\\
 \bottomrule
\end{tabular}
\caption{Existing OpenIE datasets with number of sentences, sentence source and annotation type.}
\vspace{-1em}
\label{tab:dataset_stats}
\end{table}

\paragraph{OpenIE Models.} TextRunner~\cite{yates2007textrunner} is the first OpenIE system, followed by Reverb~\cite{fader2011identifying}, OLLIE~\cite{schmitz2012open}, Clausie~\cite{del2013clausie}, CSD-IE~\cite{bast2013open}, Stanford OpenIE~\cite{angeli2015leveraging}, Openie4~\cite{mausam2016open}, Openie5\footnote{\url{github:dair-iitd/openie-standalone}}, NESTIE~\cite{bhutani2016nested}, MINIE~\cite{gashteovski2017minie} and Graphene~\cite{cetto2018graphene}. We consider them as \textit{traditional} OpenIE models. These models extract relational tuples based on handcrafted rules or statistical methods. They usually rely on prior syntactic or semantic analysis. Consequently, any error accumulated in the prior stages deteriorates model performance.

Recently, neural OpenIE systems have been developed and showed promising results~\cite{cui2018neural, zhan2020span, kolluru2020openie6, kolluru2020imojie}. Different from the traditional models, neural OpenIE models extract tuples in an end-to-end manner, not requiring prior syntactic or semantic analysis. In principle, the traditional rule-based or statistical OpenIE models do not need training. However, neural OpenIE models need a large number of training samples to learn the extraction patterns. For instance, IMOJIE~\cite{kolluru2020imojie} uses about 100,000 sentences for model training. It is unrealistic and expensive to manually annotate 100,000 sentences simply for training purpose. Therefore, a common practice in learning a neural OpenIE model is to use tuples automatically extracted by traditional systems as training data, \ie \textit{a bootstrapping strategy}. We consider these imperfect training labels generated via bootstrapping as pseudo labels. The pseudo labels used in~\cite{cui2018neural} are by Openie4, and those in~\cite{kolluru2020imojie} are from multiple OpenIE systems.

To the best of our knowledge, no OpenIE models consider document-level contexts in the tuple extraction. Nonetheless, our neural model unavoidably requires extractions of pseudo labels bootstrapped from traditional models for training. To ensure reproducibility, as part of the \dname dataset, we also release the documents that are used for generating the pseudo labels.    

%\paragraph{Boostrapping Training Labels}

%To the best of our knowledge, there is no OpenIE model performing document-level tuple extractions. Current models only process standalone sentences and extract tuples purely based on the syntactic/semantic information within this sentence. However, we argue an OpenIE model can be further improved by taking document-level context into consideration. Our proposed model \mname encodes not only the source sentence, but also its context information. Therefore \mname can be powered by document-level context.

%=============================================================================================
\section{\dname Dataset}
\label{sec:dataset}
%=============================================================================================

We now present our \textbf{Doc}ument-level context-aware \textbf{O}pen \textbf{I}nformation \textbf{E}xtraction (\textbf{\dname}) dataset. 
We first introduce the data selection and collection process, and then the annotation process by two experts. Moreover, we explain our annotation consistency measurement to indicate the high-level annotation consistency in \dname. In summary, \dname consists of two datasets: \textit{evaluation dataset} and \textit{training dataset}.
\paragraph{Evaluation dataset} contains 800 expert-annotated sentences, sampled from 80 documents in two  domains (healthcare and transportation). Specifically, 10 sentences are sampled for annotation from each of the 40 documents in one domain. In total, $2,122$ relational tuples are annotated in the 800 sampled sentences (refer to Table~\ref{tab:evaluation_stat} for detailed statistics).

\paragraph{Training dataset} contains 2,400 documents from the two domains (healthcare and transportation); 1,200 documents in each domain. All sentences from these documents are used to bootstrap pseudo labels for neural model training.\footnote{Only document IDs are included in \dname, for document collection at \url{http://patft.uspto.gov/}} 

\begin{table}
\centering
\begin{tabular}{l|rlrr}
 \toprule
  Item & \#Item & Metric & Average & Min\textapprox Max\\
 \midrule
 Doc & 80 & $N_{sent}$ & 101.78 & 44\textapprox 138\\
 \midrule
 \multirow{2}{*}{Sent} &
 \multirow{2}{*}{800} &
 $L_{sent}$ & 22.69 & 5\textapprox 47\\
 & & $N_{tuple}$ & 2.65 & 1\textapprox 8\\
 \midrule
 \multirow{3}{*}{Tuple} &
 \multirow{3}{*}{2,122} &
 $L_{sub}$ & 3.70 & 1\textapprox 17\\
 & & $L_{rel}$ & 3.39 & 1\textapprox 12\\
 & & $L_{obj}$ & 3.94 & 0\textapprox 27\\
 \bottomrule
\end{tabular}
\caption{Statistics of \dname evaluation dataset. $N_{\{\cdot\}}$ denotes the number of units (\ie sentence or tuple); $L_{\{\cdot\}}$ denotes the length (number of words) of the unit.}
\label{tab:evaluation_stat}
\end{table}

%====================================
\subsection{Dataset Collection}
\label{ssec:dcollection}
%====================================

OpenIE, by definition, is to extract relational tuples in open domain. Ideally, sentences/documents in \dname dataset shall not be restricted to any particular document type or topical domain. However, it is challenging to include all types of documents and annotate them. In fact, all existing annotations are restricted to specific types of documents like news and Wikipedia articles~\cite{niklaus-etal-2018-survey}. 

\paragraph{Document Type Selection} In building \dname, we focus on formally written documents and leave it for future work to explore other kinds of documents. We select the type of formal documents with four criteria: (i) \textbf{Adequacy:} as a document-level context-aware dataset, each document shall have a reasonable number of sentences to provide sufficient context. (ii) \textbf{Consistency:} each document shall focus on a central topic. In such a way, sentences within the same document are correlated to and consistent with each other, which helps derive proper context.  (iii) \textbf{Informativeness:} a document is considered informative if it contains informative entities like technical concepts, relations, and events. Intuitively, OpenIE models are more useful for extracting factual tuples in informative documents. (iv) \textbf{Syntactic Variety:}  sentences in these documents shall vary in syntactic structures. Such variety facilitates  thorough evaluation of OpenIE models under different scenarios.

\paragraph{Patent Document Collection} After taking all factors into consideration, we choose to collect patent documents from PatFT.\footnote{\url{http://patft.uspto.gov/netahtml/PTO/search-bool.html}} Each patent document elaborates one specific invention in reasonable length, providing sufficient contexts to annotators. They are rich in informativeness by nature, and the documents contain rich syntactic structures. 

Through PatFT search engine, patent documents can be retrieved by keywords. We have two considerations for keyword selection: 
(i) \textbf{Magnitude:} as part of \dname, a large number of documents shall be available for training  neural OpenIE models. Hence, the keywords shall lead to sufficient patent documents.
(ii) \textbf{Diversity:} the collected patent documents are expected to be diversified in  inventors, organizations, filed date, etc., to avoid fixed patterns, hence to ensure diversity of our dataset.

As the result, we choose three broad and non-technical keywords: \say{healthcare}, \say{traffic}, and \say{transportation}, to collect documents in two broad domains,  \textit{healthcare} and \textit{transportation}. Reported in Table~\ref{tab:Patprop},  42,514 and 32,256 documents are collected in healthcare and transportation respectively. Documents in each domain are contributed by more than 40,000 inventors from over 8,000 cities, and the filed dates range in several decades. 

We clean these documents by removing non-textual components in them. Then, by length (in number of words) the shortest 10\% and longest 10\%  documents are removed to avoid extremely short/long documents in our dataset. The remaining documents form the corpus from which we sample (i) documents for annotation, and (ii) training documents for bootstrapping pseudo labels.

\begin{table}[t]
\centering
\begin{tabular}{l|rr}
 \toprule
  Patents & Healthcare & Transportation\\
 \midrule
 \#Document & 42,514 & 32,256\\
 \#Inventor & 74,266 & 42,286\\
 \#Organization & 10,888 & 7,919 \\
 \#City & 10,493 & 8,095 \\
% \#State/Country  & 144 & 138 \\
 Filed year & 1999\textapprox2020 & 1970\textapprox2020\\
 \bottomrule
\end{tabular}
\caption{\label{tab:Patprop} Properties of collected patents in healthcare and transportation domains.}
\end{table}

%=======================================================================
\subsection{\dname Evaluation Dataset Selection} \label{sec:evaluation_dataset}
%=======================================================================

To ensure annotation quality and consistency, we choose to follow expert annotation scheme instead of crowdsourcing adopted in CaRB~\cite{bhardwaj2019carb}.
%Therefore, we provide only expert annotations in \dname. In this work, we rely on two OpenIE experts (authors of this paper) to perform all annotations. Due to the constraint of time and manpower, we decide to annotate 800 sentences in \dname which is comparable to CaRB (1,282 crowd-sourcing sentences) in terms of dataset size. Furthermore, comparing to the automatically constructed dataset obtained in OIE2016~\cite{stanovsky2016creating} and the crowd-sourced dataset obtained in CaRB, the expert-annotated \dname is considered more reliable and accurate in quality.
As we discussed in Section~\ref{sec:intro}, sentences exist as part of a document rather than standalone. To gain an accurate interpretation of a sentence, the annotator needs to read a few surrounding sentences, or even the entire document, for relevant contexts. Hence, the choices of labelling one sentence or multiple sentences per document incur different costs.  

%Ideally, document-level annotation should be performed on all sentences in the document in order to cover all document-level contexts. However, if we choose to annotate the whole document, we can merely cover 8 documents in our evaluation dataset because each document consists of around 100 sentences. If we solely use these 800 sentences from 8 documents for evaluation, the evaluation results of document-level OpenIE systems could be biased since only very limited OpenIE extraction patterns that can be covered by 8 documents. Therefore we target to select more documents to annotate in order to cover all-around OpenIE patterns.

To be able to cover a reasonable number of documents and also balance the annotation workload, we choose to randomly sample 10 sentences per document from 80 documents for annotation. Recall that the average number of sentences per document is 101.78 (refer to Table~\ref{tab:evaluation_stat}). The 10 sentences annotated in a document  can be used to evaluate context-aware OpenIE at 10 different positions in this document. In this way, our annotation covers 80 documents with considerable diversity.

In summary, we randomly selected 80 documents (40 in each domain) from the documents collected in Section~\ref{ssec:dcollection}. Then we randomly selected 10 sentences from each document. These 80 documents, along with 800 expert-annotated sentences form the \dname \textit{evaluation dataset}.

%====================================
%\subsection{\dname Evaluation Dataset Annotation}
%====================================

%Similar to sentence-level annotation rules mentioned in OIE2016 and CaRB, we ensured the completeness and informativeness of all tuple extractions. However, we encountered difficulties to solve both semantic and syntactic ambiguity for some sentences. Therefore we utilized their document-level contexts to determine the boundary of tuple arguments and relations. 

%To provide high quality document-level annotation, the expert annotators read through all the sentences in each document and annotated 10 randomly selected sentences of the document. By referring to the full document, the annotators could derive sufficient relevant context which is enough to resolve all the ambiguities that could not be solved at the sentence level.

%====================================
\subsection{Annotation Consistency Measurement}
%====================================
%The annotation process is particularly challenging for OpenIE task, as this task requires human annotators to fully understand both syntactic structure and semantic information prior to annotation. Moreover, as part of our document-level OpenIE annotation process, annotators are required to understand full document and apply understanding upon context to solve syntactic/semantic ambiguities. As a result, different understandings on syntactic structure, semantic meaning, or document-level context could lead to different ways of annotation. We thus measured the consistency between the two annotators.

The annotation was performed by two OpenIE experts (both are authors of this paper) with reference to existing annotation processes~\cite{stanovsky2016creating,bhardwaj2019carb}. The dataset was annotated in three stages. 

In the first stage, the two annotators practiced annotations independently on 100 sentences among the 800 sentences. Then they cross-validated the annotation results, discussed them to resolve disagreements, and updated annotation policy.

In the second stage, the two experts independently annotated another 100 sentences among the remaining 700 sentences. These two sets of annotations are used for measuring annotation consistency. Because it is not straightforward to evaluate annotation agreement by measures like Kappa coefficient, we adopted the evaluation scorer proposed by CaRB~\cite{bhardwaj2019carb}. The scorer performs matching at tuple level instead of lexical level. Specifically, we score one expert's annotations by treating the other's annotations as ground truth. Among the tuple matching strategies in CaRB, we used the default binary lenient tuple matching, to estimate the consistency between the two annotators. Reported in Table~\ref{tab:IAA}, the two annotators reach high-level agreement in annotations with an average F1 of 89.9\%. 

\begin{table}[t]
\centering
\begin{tabular}{ c|ccc}
 \toprule
 Consistency & Precision & Recall & F1\\
 \midrule
 A$\leftarrow$B & 90.7 &92.4 & 91.6 \\
 B$\leftarrow$A & 84.6 & 92.0 & 88.2\\
 \midrule
 Average & 87.7 & 92.2 & 89.9\\
 \bottomrule
\end{tabular}
\caption{Annotation consistency estimated between annotators A and B. A$\leftarrow$B indicates evaluation of A's annotations with B's annotations as ground truth.}
\label{tab:IAA}
\end{table}

Based on the high-level annotation consistency, in the third stage, each expert annotated 300 sentences from the remaining 600 sentences. The annotations are then validated by the other expert, and annotation disagreements are resolved through discussion.

%AIXIN: probably we don't need to mention it as we do not need it for now. Note that, \dname evaluation dataset does not provide co-reference information is not annotated in our work, and we leave it for future work to explore co-reference information in document-level OpenIE.

%====================================
\subsection{\dname Training Dataset}
%====================================

%\paragraph{Training Dataset} \label{para:training_dataset}
Besides the 80 documents for expert annotations, we further sample 2,400 documents randomly (1,200 in each domain) from the documents collected in Section~\ref{ssec:dcollection} to create \dname training dataset.  The 1,200 documents in each domain contain around 120,000 sentences, which is sufficient for pseudo label generation, required by neural OpenIE models.

%====================================
\section{Pseudo Label by Bootstrapping}
%====================================

%Eventually, these 2,400 documents and corresponding bootstrapped pseudo labels construct \dname \textit{training dataset}. 

% this table includes additional combination of bootstrapping strategy
\begin{table}[!t]
\centering
\begin{tabular}{ c|cccc}
 \toprule
 OpenIE Model & AUC & Prec & Rec & F1\\
 \midrule
  \multicolumn{5}{l}{Healthcare Domain} \\
 \midrule
 Reverb &35.4 & \textbf{79.9}&42.8 & 55.8\\
 Stanford & 16.5 & 11.0 & 29.7 & 16.1\\
 Clausie &22.1 & 38.8 &53.8 & 45.1\\
 OpenIE4 &35.4 & 59.5 &\underline{55.1} & 57.2\\
 OpenIE5 &29.1 & 53.6 &50.5 & 52.0\\
 Rev+Oie4 &\textbf{36.8} & \underline{75.8} &47.7 &\textbf{58.6}\\
 Oie4+Rev &\underline{35.8} & 59.6 &\textbf{55.3} & \underline{57.4}\\
 \toprule
 \multicolumn{5}{l}{Transportation Domain} \\
 %\midrule
 %System & AUC & Prec & Rec & F1\\
 \midrule
 Reverb &29.3 & \textbf{79.1}&36.3 & 49.7\\
 Stanford &15.7 & 13.2 &27.8 & 17.9 \\
 Clausie &18.0 & 36.2 &48.4 & 41.4\\
 OpenIE4 &29.2 & 52.8 &\underline{51.2} & 52.0\\
 OpenIE5 &25.0 & 50.9 &43.8 & 47.1\\
 Rev+Oie4 &\textbf{31.0} & \underline{74.2} &42.4 &\textbf{54.0}\\
 Oie4+Rev &\underline{30.1} & 53.4 &\textbf{52.7} & \underline{53.0}\\
 \bottomrule
\end{tabular}
\vspace{-0.5em}
\caption{Performance of OpenIE models on \dname evaluation dataset. The best scores are in boldface and second best scores are underlined.}
\label{tab:openie_systems}
\end{table}

Following the common practice~\cite{kolluru2020imojie,cui2018neural}, we generate pseudo labels by bootstrapping with traditional OpenIE models. Before we run these models on the \dname training dataset, we evaluate their performances on the \dname evaluation dataset, to select the models which can generate better quality pseudo labels. 

We evaluate the models by using CaRB scorer~\cite{bhardwaj2019carb}.
Table~\ref{tab:openie_systems} reports the performance of five independent OpenIE models: Reverb~\cite{fader2011identifying}, Clausie~\cite{del2013clausie}, Stanford OpenIE~\cite{angeli2015leveraging}, OpenIE4~\cite{mausam2016open} and OpenIE5\footnote{\url{github:dair-iitd/openie-standalone}}.  In addition to these five models, we also evaluated two combinations of Reverb and OpenIE4. With \textbf{Rev+Oie4}, Reverb is the main system and if Reverb fails to extract any tuples from a sentence, we complement the extraction by using Openie4. Similarly, \textbf{Oie4+ Rev} uses  OpenIE4 as the main system, and the extractions are complemented by Reverb. 

All the evaluated models show consistent performance in both domains. By F1 score, both Reverb and OpenIE4 are the best performing individual models and their combinations lead to the best and second best F1 scores in both domains. Accordingly, by applying OpenIE4, Reverb, and their combinations, the number of sentences and tuples extracted from the \dname training dataset are reported in Table~\ref{tab:psedoLabels}. Note that, a sentence is not counted if it has no extracted tuples, which leads to the different sentence number. 

\begin{table}[!t]
\centering
\begin{tabular}{p{1.6cm}|rr|rr}
 \toprule
 \multirow{2}{1.6cm}{\centering OpenIE model} &
  \multicolumn{2}{c|}{Healthcare} &
  \multicolumn{2}{c}{Transportation} \\
 & \multicolumn{1}{c}{\#Sent} & \multicolumn{1}{c|}{\#Tuple} & \multicolumn{1}{c}{\#Sent} & \multicolumn{1}{c}{\#Tuple} \\
 \midrule
  OpenIE4 & 117k & 263k & 111k & 258k\\
  Oie4+Rev & 121k & 268k & 114k & 262k\\
  Reverb & 103k & 146k & 97k & 141k\\
  Rev+Oie4 & 121k & 181k & 114k & 173k\\
  \bottomrule
\end{tabular}
\vspace{-0.5em}
\caption{Number of sentences and tuples extracted by Reverb, OpenIE4 and their combinations. The sentence is not included if it has no tuples extracted.}
\vspace{-1em}
\label{tab:psedoLabels}
\end{table}

%=============================================================================================
\section{\mname Model}
%=============================================================================================

\begin{figure}
    \centering
    \includegraphics[width=0.85\columnwidth]{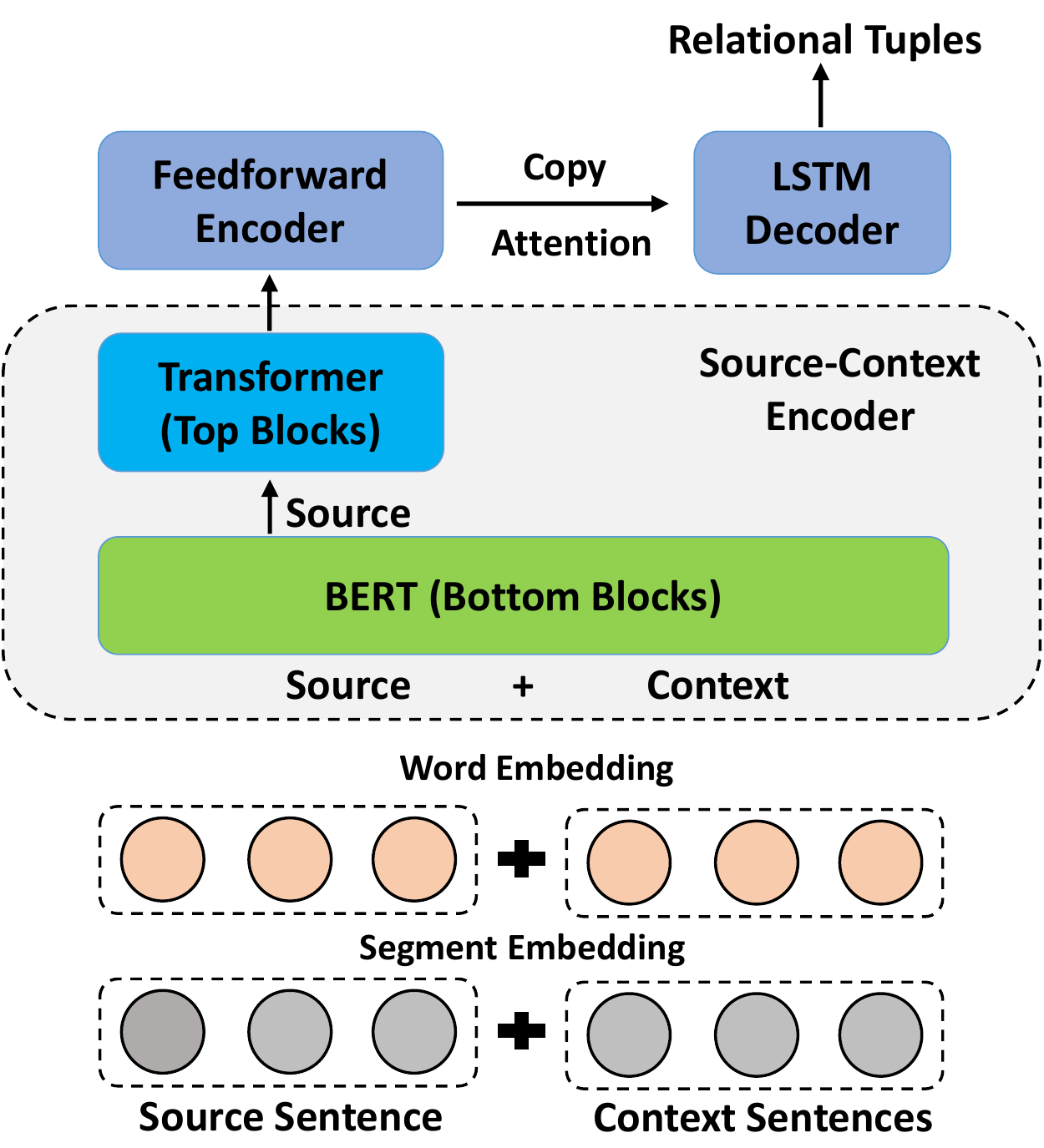}
    \caption{The architecture of \mname.}
    \vspace{-0.5em}
    \label{fig:DUCE-IE}
\end{figure}

In this section, we present the proposed \textbf{Doc}ument-level context-aware Open \textbf{I}nformation \textbf{E}xtraction model, named \textbf{\mname}. As shown in Fig.~\ref{fig:DUCE-IE}, \mname mainly consists of two parts: source-context encoder, and encoder-decoder.

\paragraph{Document-level Context} Formally, we denote a document as $D=\{s_1,s_2,\dots,s_N\}$ consisting of $N$ sentences. The source sentence $s_i$ is the input sentence that relational tuples are extracted from. Given source sentence $s_i$, we regard its surrounding sentences $c_i=\{s_{i-t},\dots,s_{i-1},s_{i+1},\dots,s_{i+t}\}$ as contextual sentences, where $t$ represents the context window size. The larger $t$ is, the more document-level context $c_i$ covers.

\paragraph{Source-Context Encoder} %We enrich the representation of source sentence $s$ with its document-level context $c$. 
The source-context encoder is inspired by a recent work~\cite{ma2020simple} which adopts Flat-Transformer to incorporate context into source sentence, for machine translation. In \mname, our encoder consists of (i) bottom blocks which take the concatenation of source sentence and context sentences as input, and (ii) top blocks which take only the representation of the source sentence from the bottom blocks as input.

In our implementation, we use BERT~\cite{devlin2018bert} as the bottom blocks to perform semantic interactions between source sentence $s$ and context $c$. We first project both $s$ and $c$ into embedding space by summing their word embedding and segment embedding, \ie $e_s=E(s)+S(s)$ and $e_c=E(c)+S(c)$. Here, $E$ is the trainable word embedding matrix, and $S$ is the trainable segment embedding matrix. The segment embedding is to distinguish words in source sentence from words in context sentences. They are initialized to 0 and 1 for words in source and context sentences respectively. Then we concatenate $e_s$ with $e_c$ as $[e_s;e_c]$ as the input to the source-context encoder.
\begin{equation}
    h_1[s;c] = \rm{BERT}([e_s;e_c])
\end{equation}
BERT, with multiple layers of transformers, merges source sentence information and its contextual information. We use the last hidden state $h_1[s;c]$ of BERT as the representation of the two concatenated input sequences.

On top of the BERT blocks, we add Transformer as top blocks~\cite{vaswani2017attention} to prepare the source sentence representation for the following encoder-decoder.
% \jjc{Just an idea you can try later: you can merge transformer top blocks and feed forward encoder into Transformer encoder and replace LSTM decoder with Transformer decoder, adding copy mechanism on top of it. in a way, you can stack a LM and a encoder-decoder. furthermore, you can use pre-trained model e.g. T5 for initialization of the encoder-decoder.}
%encode the interactions among source sentence only. 
The source sentence representation is obtained by truncating the latter (context sentences representation) $h_1[c]$ from $h_1[s;c]$. Therefore only the former (source sentence representation) $h_1[s]$ is kept.
\begin{equation}
    h_2[s] = \rm{Transformer}(h_1[s])
\end{equation}

\paragraph{Encoder-Decoder}
The encoder-decoder generation module follows CopyAttention~\cite{cui2018neural} which casts OpenIE task as a sequence-to-sequence generation task with copying mechanism. The encoder-decoder framework represents a variable length input sequence in the encoder and uses it in the decoder to generate output sequence. In our encoder-decoder framework, attention mechanism~\cite{bahdanau2014neural} is used to align the encoder hidden state with the decoder hidden state, jointly maximizing the log probability of output tuples, conditioned on the input sentence. Meanwhile, since tuple arguments and relation are normally sub-spans of the input sentence, additional copying mechanism~\cite{gu2016incorporating} is applied. It helps copy words directly from the input sentence to the output tuples.

%=============================================================================================
\section{Experiments}
\label{sec:exp}
%=============================================================================================

\begin{table*}[!t]
\centering
\begin{tabular}{ c|c|cccc|cccc}
 \toprule
 \multirow{2}{*}{Neural OpenIE} &
 \multirow{2}{*}{Pseudo labels} &
  \multicolumn{4}{c|}{Healthcare} &
  \multicolumn{4}{c}{Transportation} \\
  & & AUC & Prec & Rec & F1 & AUC & Prec & Rec & F1\\
 \midrule
 \multirow{4}{*}{CopyAttention+BERT} &
  OpenIE4 &38.6 & 54.4 & \textbf{51.6} & 52.9        &38.5 & 54.3 & \textbf{57.6} & 55.9\\
  & Oie4+Rev & 40.4 & 57.1 & 50.4 & 53.5                &38.3 & 55.3 & 56.9 & \textbf{56.1}\\
  & Reverb & 43.7 & 77.8 & 46.4 & 58.1          & 36.9 & 70.5 & 42.2 & 52.8\\
  & Rev+Oie4 & \textbf{46.8} & \textbf{77.9} & 48.6 & \textbf{59.8}     &\textbf{40.3} & \textbf{72.1} & 43.9 & 54.6\\
 \midrule
 \multirow{4}{*}{IMOJIE} &
  OpenIE4 & 36.2 & 73.0 & \textbf{47.7} & 57.7        & 35.7 & 62.9 & 48.8 & 55.0\\
  & Oie4+Rev & 34.1 & 69.5 & 46.7 & 55.9             &\textbf{35.8} & 63.5 & \textbf{49.2} & \textbf{55.5}\\
  & Reverb & 38.5 & 79.2 & 45.6 & 57.9          & 33.2 & 77.3 & 39.2 & 52.0\\
  & Rev+Oie4 & \textbf{39.7} & \textbf{80.1} & 46.4 & \textbf{58.7}     &33.0 & \textbf{77.4} & 39.6 & 52.4\\
  \bottomrule
\end{tabular}
\caption{Neural baseline models trained with different pseudo labels. The best scores of \textit{each model} are in boldface.}
\label{tab:baseline_systems}
\end{table*}

We evaluate \mname and compare its results with two baseline neural OpenIE models, CopyAttention+BERT and IMOJIE~\cite{kolluru2020imojie}. \citet{kolluru2020imojie} report that CopyAttention+BERT is a strong baseline. Meanwhile, \mname adopts CopyAttention~\cite{cui2018neural} as its encoder-decoder module. Hence CopyAttention+BERT can be considered as the base model, from which \mname adds context modelling.  

%=================================
\subsection{Neural Baseline Models}
%=================================
We first evaluate the two neural baseline models trained with the pseudo labels listed in Table~\ref{tab:psedoLabels}. The evaluation is conducted on the \dname evaluation dataset with CaRB scorer.

Reported in Table~\ref{tab:baseline_systems}, CopyAttention+BERT outperforms IMOJIE in most settings by both measures: AUC and F1. In general, for both models, pseudo labels by Rev+Oie4 (and also Reverb) lead to better results in healthcare domain. Pseudo labels by Oie4+Rev (and also OpenIE4) generate better results in transportation domain.  During our annotation of the 800 sentences,  we observe that sentences in transportation domain tend to contain slightly more conjunctions (\eg \say{and} and \say{or}) and thus have more coordinating structures than those in healthcare. OpenIE4 system generally extracts more tuples than Reverb (refer to Table~\ref{tab:psedoLabels}) and provides higher recall. Therefore, extractions in transportation domain with more conjunctions may better match the tuples extracted by OpenIE4.

Based on this set of results, in our following experiments, we use pseudo labels by Rev+Oie4 for healthcare domain, and pseudo labels by Oie4+Rev for transportation domain.

%Hence we conclude that Rev+Oie4 is the most suitable bootstrapping training data in Healthcare domain. On the other hand, in Transportation domain, baseline models trained with Oie4+Rev bootsrapping pseudo labels achieve best AUC and F1.
%In our \dname evaluation dataset, 400 Healthcare sentences have 921 relational tuples annotated while 400 Transportation sentences have 1201 tuples.
%The reason for this discrepancy might be due to different writing styles in the two domains. By observation, sentences in transportation domain tend to contain more conjunctions (e.g., `and' and `or') and thus have more coordinating structures than those in healthcare. Conjunctions lead some OpenIE systems to more tuple extractions from a sentence. For instance, Openie4 system generally extracts more tuples than Reverb, thus achieving better recall but lower precision than Reverb. As a result, Transportation domain with more conjunctions favors Openie4 that can provide better recall for it.

%====================================
\subsection{\mname Against Baselines}\label{sec:results}
%====================================

\begin{table*}
\centering
\begin{tabular}{ c|cccc|cccc}
 \toprule
 \multirow{2}{*}{System} &
  \multicolumn{4}{c|}{Healthcare} &
  \multicolumn{4}{c}{Transportation} \\
  & AUC & Prec & Rec & F1 & AUC & Prec & Rec & F1\\
 \midrule

 Rev+Oie4 & 36.8 & 75.8 &47.7 & 58.6          & 31.0 & \textbf{74.2} &42.4 & 54.0\\
 Oie4+Rev &35.8 & 59.6 & \textbf{55.3} & 57.4     &30.1 & 53.4 & 52.7 & 53.0\\
 \midrule
 CopyAttention+BERT &  46.8 & \underline{77.9} & 48.6 & 59.8           &\underline{38.3} & 55.3 & 56.9 & 56.1\\
 IMOJIE & 39.7 & \textbf{80.1} & 46.4 & 58.7           &35.8 & \underline{63.5} & 49.2 & 55.5 \\
 \midrule
 \mname w/o transformer & \underline{47.1} & 76.2 & 49.9 & \underline{60.3}           & \textbf{38.5} & 55.8 & \underline{57.0} & \underline{56.4}  \\
 \mname w transformer & \textbf{47.4} & 74.4 & \underline{51.3} & \textbf{60.8}           & \textbf{38.5} & 56.0  & \textbf{57.5} & \textbf{56.9} \\
 \bottomrule
\end{tabular}
\vspace{-0.6em}
\caption{Results of \mname and baselines. The best scores are in boldface and second best scores are underlined.}
\vspace{-0.6em}
\label{tab:final_results}
\end{table*}

In this section, we evaluate \mname against sentence-level OpenIE systems. We refer \mname without the top transformer layer as \say{\mname w/o transformer} and \mname as \say{\mname w transformer} for clarity. The context window size of \mname is set to 5 for healthcare domain and 4 for transportation domain. 

Table~\ref{tab:final_results} summarizes the experiment results. For easy comparison, the results of the best traditional OpenIE baselines (refer to Table~\ref{tab:openie_systems}) and neural OpenIE models (refer to Table~\ref{tab:baseline_systems}) are replicated here. Observe that \mname w transformer achieves the best AUC and F1 in both domains. Its variant, \mname w/o top transformer, is the second best performer and outperforms all sentence-level models. 

The experiment results suggest that incorporating document-level context is helpful in improving OpenIE. On the other hand, we remark that \mname is trained by pseudo labels produced by traditional OpenIE models which do not consider document-level context. The potential of utilizing document-level context is yet to be fully realized.

%====================================
\subsection{Impact of Context Window Size}\label{sec:results_window}
\begin{figure}[t]
    \centering
	\subfigure[Healthcare]
	{\label{sfig:window_health}	\includegraphics[width=0.23\textwidth]{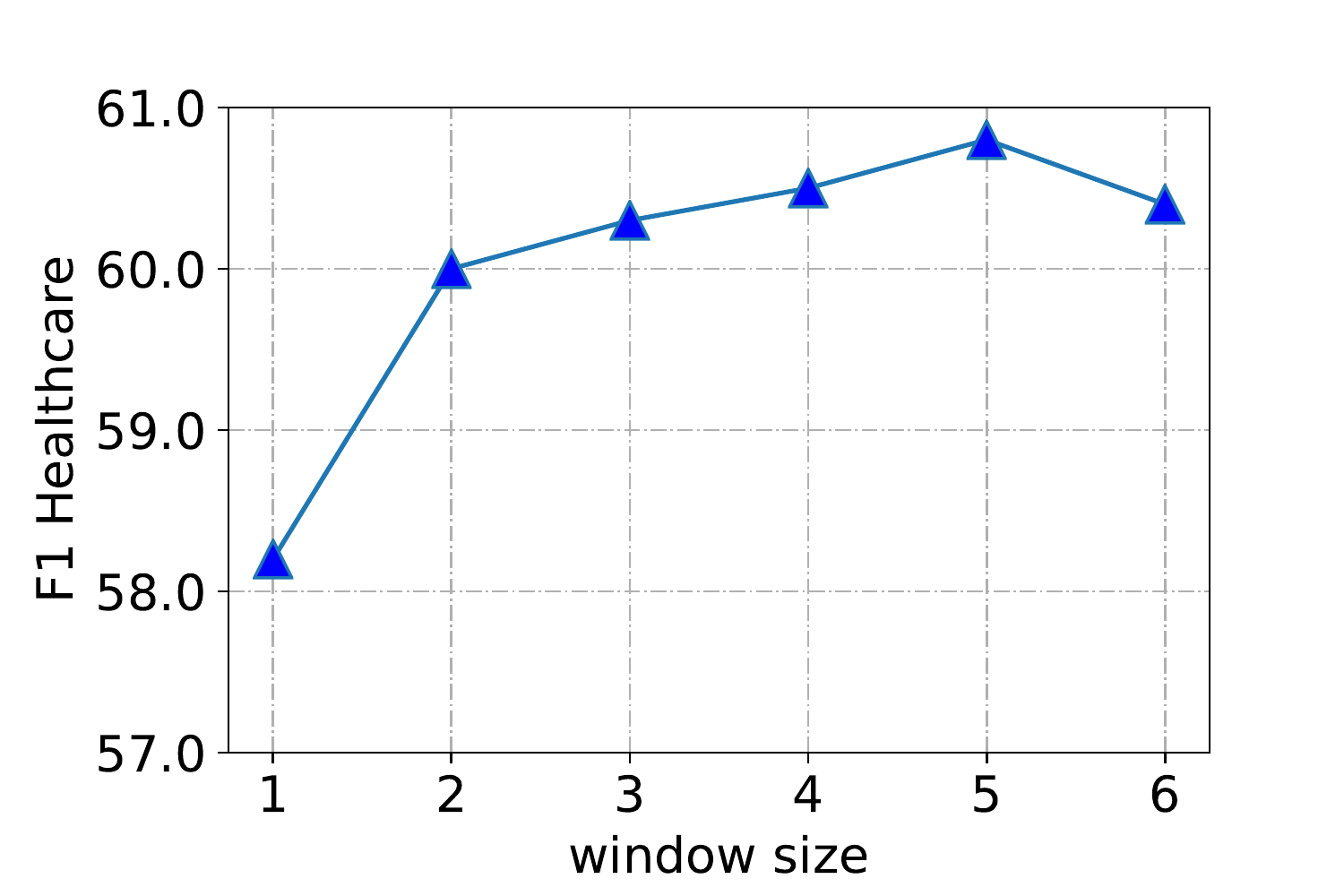}}
	\subfigure[Transportation]
	{\label{sfig:window_trans}	\includegraphics[width=0.23\textwidth]{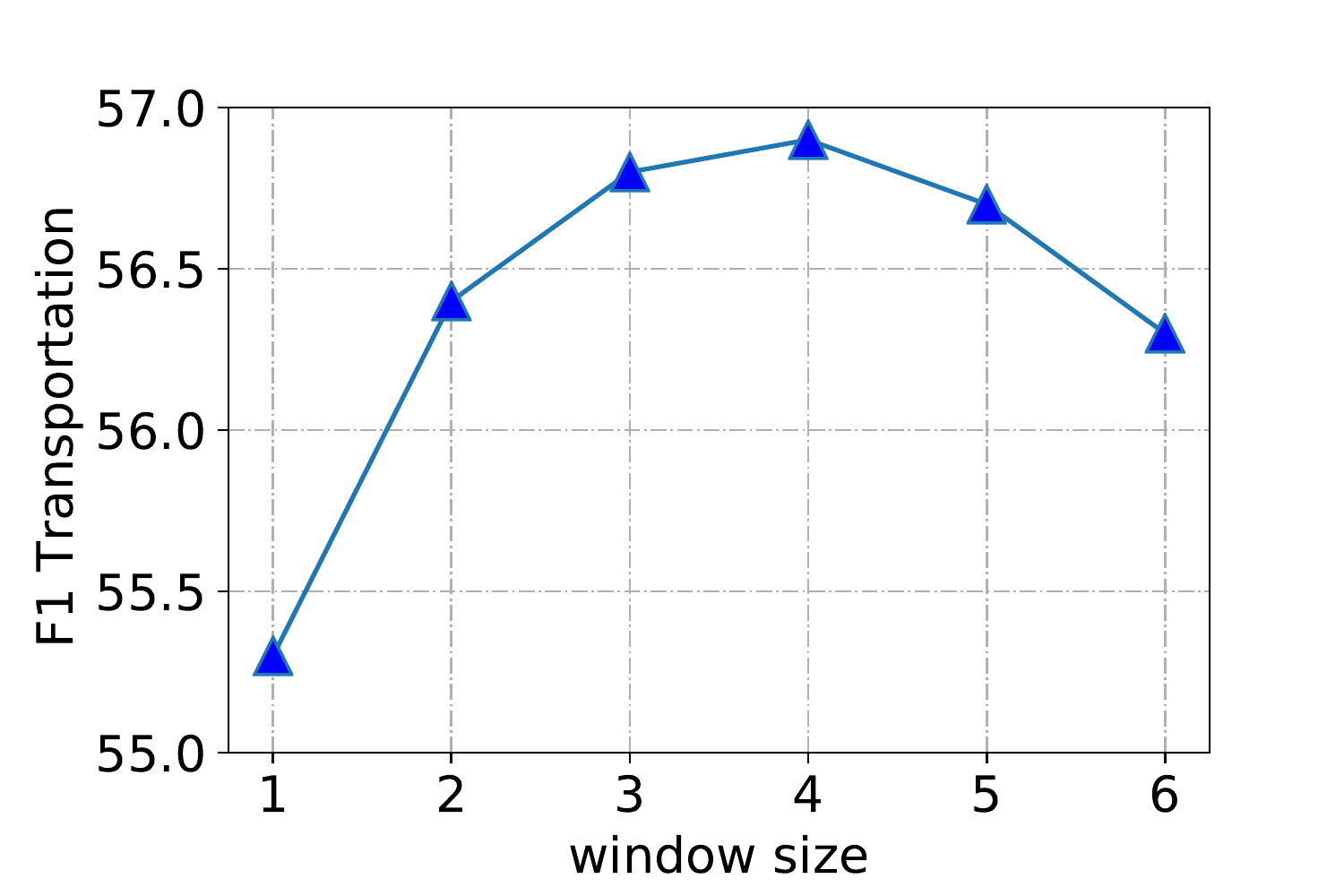}}
	\vspace{-1em}
	\caption{F1 with varying window sizes.}
	\vspace{-0.8em}
	\label{fig:win_size}
\end{figure}

%====================================
The setting of window size determines the number of context sentences to be considered. We evaluated the range from 1 to 6 and plot F1 scores against window size changes in Fig.~\ref{fig:win_size}. Observe that the optimal window size for healthcare domain is 5, and the number is 4 for transportation. Better F1 scores are observed along the increase of context sentence window, till 4 or 5. In general,  8 \textapprox 10 surrounding sentences (window size 4 or 5) provide sufficient context for sentence understanding. Small window size might not provide sufficient context, and a large window size might introduce noise and dominate source representation learning.

%We assume the longer sequence of context taken, the richer context information our model can benefit. However, pretrained BERT block limits our source-context length to 512 tokens. We need to consider the trade-off between size of window and length of each context sentence. The average length of the sentences is 22.69 tokens as shown in Table.~\ref{tab:dataset_stats}. Hence we limit the length of each context sentence to be 40 and the sentence exceeding this limit will be truncated. Finally we conduct our experiment on different window sizes ranging from 1 to 6 to explore the relation between context information length and model performance. The results of F1 and AUC of DocIE against window size is shown in Fig.~\ref{fig:win_size}. We observe the optimal window size in healthcare domain is 5, at which both F1 and AUC are achieved. In transportation domain, F1 achieves highest score when window size equals to 4. Hence we conclude that 8 \textapprox 10 surrounding sentences (window size 4 or 5) could provide optimal context for \mname to leverage. Smaller window size might not provide efficient information to derive context. And larger window size could distract the original meaning of source sentence.

%====================================
\subsection{Case Study}
%====================================

We use the two example sentences shown in Fig.~\ref{fig:example} as a case study, to illustrate the differences between \mname and the sentence-level neural OpenIE baselines: CopyAttention+BERT and IMOJIE.

For Sentence 1, CopyAttention+BERT incorrectly recognizes the word \say{transfers} as a verb, thus extracting an incorrect tuple (data ; transfers to ; a single target terminal). IMOJIE, however, completely misses the key phrase \say{data transfers} and extracts an incorrect tuple (a single target terminal ; using ; the invention). Only \mname manages to extract the correct tuple (data transfers to a single target terminal ; using ; the invention).

For Sentence 2, there is no explicit clue about the relationship between \say{a device} and \say{a cellular base}. Both CopyAttention+BERT and IMOJIE treat \say{a device a cellular base station} as a whole and
mistakenly generate a tuple (Node-B ; can be ; a device a cellular base station having beam-forming antennas). In contrast, \mname successfully splits \say{a device a cellular base station} by referring to surrounding context and extracts the correct tuple (Node-B ; can be ; a device). However, \mname fails to infer the inter-relationship between \say{a device} and \say{a cellular base}. Accordingly, another correct tuple (a device ; is such as ; a cellular base station) is not extracted.

Results of the two example sentences show the improvements made by \mname after leveraging contextual information for tuple extraction.

%==========================================
\subsection{Error Analysis}
%==========================================

Similar to the error analysis performed in~\cite{kolluru2020imojie}, we examine tuples extracted by \mname from 50 randomly selected sentences in \dname. We identify the following major error types. (i) \textbf{Incompleteness:} In 28\% sentences, \mname fails to cover at least one key phrase in either arguments or relation. Missing key phrases result in incomplete information extraction. (ii) \textbf{Incorrect Boundary:} 27\% extractions misinterpret the syntactic meaning of the sentence, leading to incorrect boundary of arguments and relation. (iii) \textbf{Redundant Extractions:} 15\% sentences contain redundant extractions; that is, the same relational fact is extracted multiple times from a sentence or phrase. (iv) \textbf{Grammatical Errors:} 13\% extractions are not grammatically correct. Most  grammatical errors are contributed by the incorrect verb form used in tuple relation.

%====================================
\subsection{Implementation}
%====================================
We implement \mname using the AllenNLP framework\footnote{\url{https://github.com/allenai/allennlp}} in Pytorch 1.4. Pre-trained BERT\footnote{\url{https://huggingface.co/transformers/model_doc/bert.html}} is fine-tuned at learning rate $2 \times 10^{-5}$ to get contextualized word embeddings. The learning rate for the other modules is set to $1 \times 10^{-4}$. The input dimension, projection dimension, feedforward hidden dimension, number of layers, and number of attention heads of top transformer encoder are set to 768, 256, 3072, 2, and 8, respectively. The hidden dimension, and word embedding dimension of the LSTM-decoder are set to 256 and 100 respectively.

%=====================================================
\section{Conclusion}
%=====================================================

In this research, we propose to consider document-level contextual information for OpenIE task. We contribute \dname, the first document-level context-aware OpenIE dataset. It consists of 800 expert-annotated sentences from 80 documents. The documents are carefully selected and the annotations are completed by experts with high-level annotation consistency. 

With the help of \dname, we conduct evaluation of neural OpenIE models and demonstrate that incorporating document-level context is helpful in improving OpenIE performance through \mname.  As a baseline for document-level context-aware OpenIE, \mname achieves promising results compared with all sentence-level OpenIE models. Our future works are in two main directions. One is to research on more effective context-aware OpenIE models, and the other is to investigate the possibility of not relying on pseudo labels.

\bibliographystyle{acl_natbib}
\bibliography{acl2021}

\end{document}